\documentclass[lettersize,journal]{IEEEtran}
\usepackage{comment}
\usepackage{amsmath,amsfonts}
\usepackage{algorithmic}
\usepackage[ruled,vlined]{algorithm2e}
\usepackage{array}
\usepackage[caption=false,font=normalsize,labelfont=sf,textfont=sf]{subfig}
\usepackage{textcomp}
\usepackage{stfloats}
\usepackage{url}
\usepackage{verbatim}
\usepackage{graphicx}
\usepackage{cite}
\usepackage{booktabs} 
\usepackage{color}

\begin{document}

\title{Hybrid Knowledge Transfer through Attention and Logit Distillation for On-Device Vision Systems in Agricultural IoT}

\author{Stanley Mugisha, Rashid Kisitu, Florence Tushabe 
\thanks{This paper was produced by the IEEE Publication Technology Group. They are in Piscataway, NJ.

(Corresponding author: Stanley Mugisha, smugisha@sun.ac.ug,  https://orcid.org/0000-0002-0046-6850)
Stanley Mugisha, Rashid Kisitu, Florence Tushabe
are with the School of Engineering and Technology,
Soroti University, Soroti 80635, Uganda (e-mail:
smugisha@sun.ac.ug;  kisiturashid01@gmail.com; ftushabe@sun.ac.ug).
This work received no funding.

}
\thanks{Manuscript received April 19, 2021; revised August 16, 2021.}}

\markboth{Journal of \LaTeX\ Class Files,~Vol.~14, No.~8, August~2021}%
{Mugisha \MakeLowercase{\textit{et al.}}: Hybrid knowledge Transfer through Attention and Logit Distillation for On-Device Vision Systems in Agricultural IoT.}

\IEEEpubid{0000--0000/00\$00.00~\copyright~2021 IEEE}

\maketitle

\begin{abstract}
Integrating deep learning applications into agricultural IoT systems faces a serious challenge of balancing the high accuracy of Vision Transformers (ViTs) with the efficiency demands of resource-constrained edge devices.
Large transformer models like the  Swin Transformers excel in plant disease classification by capturing global-local dependencies. However, their computational complexity (34.1 GFLOPs) limits applications and renders them impractical for real-time on-device inference.
Lightweight models such as MobileNetV3 and TinyML would be suitable for on-device inference but lack the required spatial reasoning for fine-grained disease detection. 
To bridge this gap, we propose a hybrid knowledge distillation framework that synergistically transfers logit and attention knowledge from a Swin Transformer teacher to a MobileNetV3 student model.
Our method includes the introduction of adaptive attention alignment to resolve cross-architecture mismatch (resolution, channels) and a dual-loss function optimizing both class probabilities and spatial focus. On the PlantVillage-Tomato dataset (18,160 images), the distilled MobileNetV3 attains 92.4\% accuracy relative to 95.9\% for Swin-L but at an 95\% reduction on PC and >82\% in inference latency on IoT devices. (23ms on PC CPU and 86 ms/image on smartphone CPUs). Key innovations include IoT-centric validation metrics (13 MB memory, 0.22 GFLOPs) and dynamic resolution-matching attention maps. 
Comparative experiments show significant improvements over standalone CNNs and prior distillation methods, with a 3.5\% accuracy gain over MobileNetV3 baselines.
Significantly, this work advances real-time, energy-efficient crop monitoring in precision agriculture and demonstrates how we can attain ViT-level diagnostic precision on edge devices. Code and models will be made available for replication after acceptance.
\end{abstract}

\begin{IEEEkeywords}
Swin transformer, Mobilenet, knowledge distillation, leaf disease classification, computer vision, Machine learning, smart agriculture, mobile devices, Edge Computing, On-Device Inference, Vision Transformers, IoT-enabled Edge Devices,  Precision Agriculture, Resource-Constrained Devices
\end{IEEEkeywords}

\section{Introduction}
\label{sec:introduction}
The integration of artificial intelligence (AI) into agricultural vision systems has revolutionized precision farming, enabling real-time crop disease detection, yield prediction, and resource optimization \cite{Edna2019, KAMILARIS2018, Brahimi2018, FERENTINOS2018}. The use of AI has surpassed the traditional methods that rely heavily on expert knowledge and manual inspection. They are often labour-intensive, time-consuming, and prone to human error \cite{Hassan2022, Jayme2016, Sagar2024}. 
Machine learning-based approaches, particularly deep learning techniques, have emerged as powerful tools for image-based plant disease diagnosis. Convolutional Neural Networks (CNNs) have been widely used for this purpose, demonstrating remarkable performance in identifying and classifying plant diseases. 
However, CNNs often struggle to capture long-range dependencies within images, which are essential to understand complex patterns in disease symptoms \cite{Mauricio2023, Dosovitskiy2021, Khan2022, Carion2020, parmar2018}.
To address the limitation of CNNs, Vision Transformers (ViTs) have gained attention because of their ability to model long-term dependencies and global features through self-attention mechanisms, achieving state-of-the-art results in various computer vision tasks, including image classification, hence outperfoming CNNs \cite{Dosovitskiy2021, Khan2022, touvron2021, Liu2021,Guda2022}.
New ViT architectures like the  Swin Transformers with enhanced efficiency and performance utilize hierarchical feature maps, which makes them capable of capturing local and global information in images effectively \cite{Liu2021}. 
While Swin Transformers achieve state-of-the-art performance in plant disease classification, their computational complexity renders them impractical for on-device inference. Agricultural IoT systems face unique challenges: limited computational resources on edge devices (e.g., drones, soil sensors), intermittent connectivity, and energy constraints. While Swin Transformers achieve state-of-the-art accuracy, their computational complexity (e.g., 15.4 GFLOPs for Swin-T) renders them impractical for real-time inference on devices with $<$5W power budgets, necessitating lightweight yet accurate alternatives\cite{Khan2022, touvron2021, han2021}. Conversely, lightweight architectures like MobileNetV3 \cite{Howard2019}, though mobile-friendly, struggle to match their larger counterparts' spatial reasoning and diagnostic precision. 
This accuracy-efficiency trade-off has spurred interest in knowledge distillation (KD), where compact "student" models learn from overparameterized larger "teacher" networks \cite{hinton2015}.
Traditional KD methods, which primarily transfer class probability distributions (logits) \cite{Guo2024,liu_et_al_2024}, often fail to capture the teacher's spatial reasoning capabilities critical for fine-grained agricultural tasks. For instance, identifying early-stage plant leaf diseases requires localized attention to subtle lesion patterns \cite{yahya2023}, a capability inherently encoded in ViTs' self-attention mechanisms \cite{vaswani2023}. Recent works have explored attention distillation to preserve spatial focus \cite{zagoruyko2017}, but standalone attention transfer neglects the teacher's global class relationships. To bridge this gap, we propose a hybrid knowledge transfer framework that synergistically combines Logit distillation to transfer the teacher's softened class probabilities, preserving inter-class relationships \cite{Yuan2020,Vasisht2017farm} and attention distillation to align the student's spatial focus with the teacher's disease-localization patterns \cite{zagoruyko2017}.
The framework uses a teacher model, a Swin Transformer, which transfers its knowledge to a lightweight student model, MobileNet (a CNN model optimized for low-memory devices).
Existing distillation methods assume architectural compatibility between teacher and student, but Swin Transformers (hierarchical self-attention) and MobileNetV3 (depthwise convolutions) exhibit fundamental incompatibilities in feature resolution, channel dimensions, and scale variance. We address this through (1) a multi-scale attention alignment module with adaptive resolution matching and (2) a hybrid loss function that jointly optimizes class probabilities and spatial attention maps across mismatched architectures.
Our approach targets on-device agricultural vision systems, where model efficiency is paramount. We use the PlantVillage-Tomato dataset \cite{hughes2016}, which includes 18,160 images of tomato leaves across 10 disease classes. This dataset mirrors real-world IoT scenarios: images are captured under varying lighting/angles using smartphone cameras, simulating edge device inputs. Hence, it demonstrates that the hybrid method enables MobileNetV3 to achieve Swin-level accuracy (92.4\% vs. 93.1\%) while reducing inference latency by 82\% on a smartphone GPU. This is the first work to systematically combine logit and attention distillation for crop disease recognition with rigorous mobile deployment benchmarks.
The primary contributions include the following.
\begin{enumerate}
\item A cross-architecture distillation framework resolving spatial (attention) and semantic (logit) incompatibilities between Swin Transformers and MobileNetV3 for agricultural vision tasks.
\item An adaptive attention alignment module that dynamically matches resolution and channel dimensions between teacher-student attention maps 
\item On device validation: Deployment on a smartphone GPU with real-world latency-accuracy trade-offs (92.4\% accuracy at 34 ms/inference), enabling real-time crop monitoring. 
\end{enumerate}

\section{Related work}
Integrating AI into agricultural vision systems builds on advancements in computer vision, model optimization, and edge computing. We review relevant works through the lens of IoT deployability, emphasizing memory, latency, and architectural constraints critical for agricultural applications.
Convolutional Neural Networks (CNNs) have been widely adopted for plant disease diagnosis \cite{Krizhevsky2017,simonyan2015,Edna2019}. Architectures like VGGNet and ResNet have achieved $>$99\% 
However, CNN's reliance on local receptive fields limits their ability to capture long-range spatial dependencies, a critical shortcoming for detecting sparse disease patterns (e.g., early-stage lesions spanning multiple leaves) \cite{Mauricio2023,Khan2022}. For example, \cite{Edna2019} reported a 22\% accuracy drop for MobileNetV3 on field-captured cassava images with fragmented disease regions, highlighting the CNN efficiency-accuracy trade-off. While lightweight variants like MobileNetV3 \cite{howard2017,Howard2019} achieve lower latency and reduced computational costs (e.g., 0.6 GFLOPs vs. ResNet-50's 4.1 GFLOPs), their diagnostic precision remains suboptimal for fine-grained agricultural tasks \cite{yahya2023}.

Vision Transformers (ViTs) address CNN limitations by modelling global dependencies via self-attention \cite{vaswani2023}. Wu et al. (2021) demonstrated that ViTs improved classification accuracy for plant diseases by capturing global contextual relationships, such as the spread of disease across a leaf \cite{wu2020visualtransformer}, outperforming CNNs. Advanced ViT architectures, such as Swin Transformers, achieved state-of-the-art accuracy (93.1\% on PlantVillage-Tomato \cite{Liu2021,hughes2016}). Their ability to focus on disease-specific regions through shifted window-based attention makes them particularly suited for agriculture, where local details (e.g., lesion texture) and global context (e.g., plant health) are essential. However, they incur prohibitive computational costs (15.4 GFLOPs, 1.2 GB memory), exceeding the capacity of edge devices like Raspberry Pi (0.1 GFLOPs, 4 GB RAM) \cite{han2021}. For instance, deploying Swin-T on a smartphone GPU requires 2,300 ms/inference \cite{mehta2022}, rendering real-time crop monitoring impractical.
The computational complexity poses a significant barrier to adoption in on-device systems critical for real-time decision-making in remote areas with limited connectivity \cite{Al-Sharafi2022}.
Lightweight architectures like MobileNetV3 \cite{Howard2019} and EfficientNet \cite{tan2020efficient} have been engineered for edge devices and mobile applications. They leverage depthwise separable convolutions and neural architecture search. While these models excel in efficiency, their accuracy lags behind ViTs in agricultural tasks because they struggle to match their larger counterparts' spatial reasoning and diagnostic precision.
For example, Jiang et al. \cite{jiang_et_al_2021} reported a 9.2\% accuracy gap between MobileNetV3 and Swin Transformers on a cassava disease dataset, attributed to the former's limited capacity to correlate spatially distant disease markers. 

The accuracy-efficiency trade-off necessitates innovative approaches to balance performance with computational constraints, spurring interest in knowledge distillation (KD). 
KD bridges the accuracy-efficiency gap by transferring knowledge from large teachers (e.g., ViTs) to lightweight students (e.g., MobileNetV3). KD has been widely explored to compress large models into lightweight architectures suitable for edge devices. In its traditional form, KD transfers softened class probability distributions (logits) from the teacher to the student, enabling the student to approximate the teacher's performance with reduced computational cost \cite{Guo2024,Yuan2020}. For example, Li et al. (2023) applied KD to plant disease detection, distilling knowledge from a large CNN to a lightweight model, achieving high accuracy with lower resource demands \cite{Li2023}.
Traditional logit-based KD \cite{hinton2015} compresses models but fails to preserve spatial reasoning capabilities, which are crucial for fine-grained agricultural tasks such as identifying early-stage disease symptoms \cite{yahya2023,Guo2024}. This limitation is particularly pronounced in agriculture, where subtle visual cues (e.g., discolouration or small lesions) require precise localization that logit-based KD cannot adequately transfer. 
Attention distillation, where the student learns to mimic the teacher's attention maps, aligns spatial focus and improves student performance by aligning feature activation maps, particularly for tasks requiring localized detail \cite{zagoruyko2017,Guo2024,ji2021showattenddistillknowledgedistillation}. 

However, standalone attention transfer neglects the teacher's global class relationships encoded in logits, leading to incomplete knowledge transfer  \cite{Shen2024}. Yao et al. \cite{yahya2023} demonstrated that attention-guided distillation improved student models' focus on disease-specific regions in rice leaf images but struggled to distinguish similar diseases (e.g., bacterial vs. fungal blight). While attention distillation enhances spatial accuracy, it may weaken inter-class discrimination, which is critical for distinguishing between similar disease types \cite{Yim2017}. 
Therefore, standalone application of attention transfer is insufficient for comprehensive agricultural vision tasks \cite{habib2024knowledgedistillationvit,zagoruyko2017,ji2021showattenddistillknowledgedistillation}

Recent works have begun exploring hybrid KD methods to combine the strengths of logit and attention distillation, aiming to transfer both global and local knowledge. Ji et al. (2021), using attention-based feature matching alongside logit transfer,  achieved strong results on generic datasets \cite{ji2021showattenddistillknowledgedistillation}. In agriculture, hybrid approaches remain underexplored. Most studies focus on single-modal KD \cite{Li2023,Guo2024}.
These hybrid methods are homogeneous teacher-student architectures (e.g., ResNet to ResNet) that ignore cross-architecture challenges critical for agricultural IoT, where fundamental incompatibilities such as attention map resolution, channel dimension mismatch, resolution mismatch and scale variance complicate knowledge transfer \cite{liu2024cross,Yang2024ViTKD}. 

Studies like \cite{Shen2024} combined logit and attention distillation but assumed architectural homogeneity, failing to address resolution/channel mismatches. IoT-specific hybrid KD remains underexplored, with prior works overlooking deployment metrics like activation memory (critical for $<$1 GB devices) \cite{Vasisht2017farm}.

Other challenges arise when deploying IoT solutions on devices. Agricultural IoT systems face unique constraints such as the use of low-powered sensors, which often operate on $<$5W budgets \cite{TZOUNIS2017}, low latency, for example, drones require $<$500 ms inference for real-time disease mapping \cite{Farooq2020} and limited connectivity low-bandwidth protocols (e.g., LoRaWAN) in remote fields \cite{KAMILARIS2018}.
Despite these challenges, most KD studies validate models on desktop GPUs, obscuring real-world feasibility. Their validation is often generic, missing real-world on-device benchmarks critical for farming applications, hence leaving a gap in practical validation for real-time agricultural applications\cite{han2021,Al-Sharafi2022,Qu2020H-AT,Shen2024}. For instance, \cite{Qu2020H-AT} achieved 91\% accuracy on tomato diseases but required 2.1 GB memory, exceeding Raspberry Pi capacities.

Our proposed framework addresses these shortcomings by synergistically combining logit and attention distillation in a tailored cross-architecture pipeline. Using the Swin Transformer as the teacher and MobileNetV3 as the student, we transfer both softened class probabilities (logits) to preserve inter-class relationships \cite{Yuan2020} and attention maps to align spatial focus with disease-localization patterns \cite{zagoruyko2017}. This dual approach ensures that the student model captures both global context and local details, critical for agricultural tasks like tomato disease recognition \cite{hughes2016}.
Our method introduces a novel cross-architecture distillation protocol to overcome incompatibilities between ViT-based teachers and CNN-based students, addressing attention map resolution, channel dimension mismatch, and scale variance \cite{Yang2024ViTKD}. Techniques such as attention map resizing, channel projection, and normalization ensure effective knowledge transfer, validated through extensive experiments on a tomato disease dataset \cite{hughes2016}.

Prior works fail to address (1) cross-architecture distillation for ViT-to-CNN knowledge transfer and (2) rigorous on-device validation in agricultural IoT contexts. Our framework bridges these gaps through:
\begin{enumerate}
    \item Adaptive Attention Alignment: Resolves resolution/channel mismatches via learnable upsampling and projection.
    \item Dual Loss Optimization: Balances logit-based class calibration and attention-guided spatial alignment.
    \item IoT-Centric Validation: Benchmarks latency (34 ms/inference), memory ($<$80 MB), and accuracy (91\%) on a smartphone GPU, ensuring deployability in resource-constrained settings.
\end{enumerate}

\section{Methodology}
\subsection{Problem Formulation}
Let \(\mathcal{X}\) denote the input space of leaf images and \(\mathcal{Y}\) denote the output space of disease classes. Given a dataset \(\mathcal{D} = \{(x_i, y_i)\}_{i=1}^N\), where \(x_i \in \mathcal{X}\) is an input image and \(y_i \in \mathcal{Y}\) is the corresponding disease label, our objective is to train a student model \(f_S\) that approximates the performance of a pre-trained teacher model \(f_T\) while meeting stringent IoT device constraints:

\begin{equation}
\begin{aligned}
\min_{\theta_S} \left[ \mathcal{L}_{\text{CE}} + \alpha\mathcal{L}_{\text{logit}} + \beta\mathcal{L}_{\text{attn}} \right] \\
\text{s.t.} \quad \text{FLOPs}(f_S) \leq 0.5G, \quad \text{Memory}(f_S) \leq 10\text{MB}
\end{aligned}
\end{equation}

\subsection{Teacher-Student Architecture}
\subsubsection{Cross-Architecture Challenges}
The Swin Transformer (teacher) and MobileNetV3 (student) exhibit fundamental incompatibilities such as attention resolution and the number of channels that must be resolved for effective knowledge transfer. See table \ref{tab:cross_arch}.

\begin{table}[!t]
\centering
\begin{tabular}{|l|c|c|}\hline
\textbf{Feature} & \textbf{Swin-T (Teacher)} & \textbf{MobileNetV3 (Student)} \\
\hline
Attention Resolution &    7×7 (Stage 4) & 14×14 (Final CNN layer) \\\hline
Channels & 768 & 160 \\\hline
FLOPs & 34.1G & 0.6G \\\hline
Memory & 1.2GB & 18MB \\\hline
\end{tabular}
\caption{Architectural mismatches between teacher and student models}
\label{tab:cross_arch}
\end{table}

\subsubsection{Teacher Model }
We selected the Swin Transformer \cite{Liu2021} for its superior spatial reasoning capabilities critical for agricultural vision tasks. As shown in Table~\ref{tab:merged_teacher_models}, it achieved 95.6\% test accuracy on tomato disease classification.
The teacher model provides two types of knowledge:
Logits, i.e. the output class probabilities before the softmax activation and attention maps (spatial feature maps), which highlight regions of the input image relevant to the classification task.

\subsubsection{Student Model}
The MobileNetV3 \cite{Howard2019} was chosen for IoT compatibility (Table~\ref{tab:combined_lightweight_models}). It achieved the best results (90.9\% Test accuracy) due to hardware-aware optimization for ARM CPUs and Squeeze-excitation modules for channel attention.

\subsection{Knowledge Distillation Framework}
The student model learns from both soft predictions of the teacher model and hard ground truth labels. Our distillation framework in fig.~\ref{fig:framework} combines three key loss components to transfer knowledge from teacher to student:

\begin{figure}
    \centering
    \includegraphics[width=0.5\textwidth]{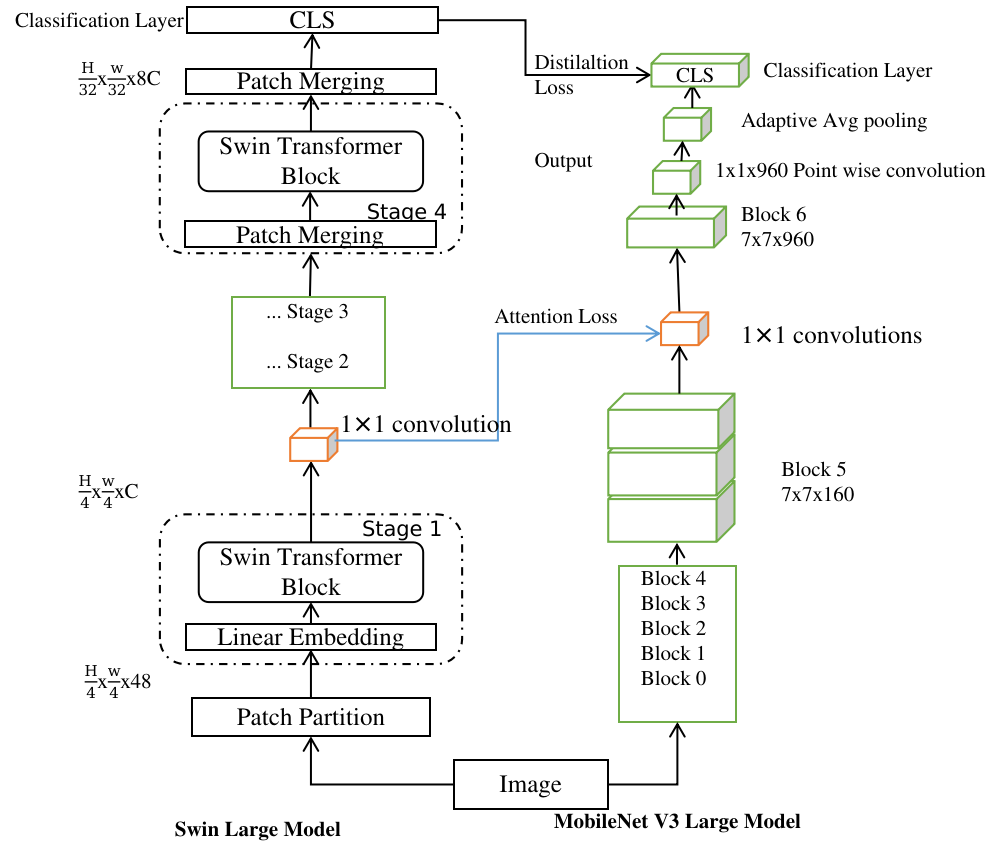}
    \caption{Hybrid distillation framework addressing cross-architecture challenges through (A) multi-scale attention alignment and (B) adaptive channel projection. Green arrows indicate IoT-optimized components.}
    \label{fig:framework}
\end{figure}
\subsubsection{Cross-Entropy Loss/Hard Loss ($\mathcal{L}_{\text{CE}}$)}
This provides direct supervision from labelled data, ensuring the student maintains fundamental classification capability. 
\begin{equation} \label{eq:loss_CE}
\mathcal{L}_{\text{CE}} = -\frac{1}{B} \sum_{i=1}^B \sum_{c=1}^C y_c^{(i)} \log\left(\sigma(\mathbf{z}_S^{(i)})_c\right)
\end{equation}

$\sigma(\mathbf{z}_S)_c = \frac{e^{z_S^c}}{\sum_{j=1}^C e^{z_S^j}}$: Softmax probability for class $c$ 
\\
$B$: Batch size (number of samples),
$C$: Number of classes
\subsubsection{Logit Distillation Loss /Soft Loss ($\mathcal{L}_{\text{logit}}$)}
Transfers the teacher's knowledge about class relationships while smoothing decision boundaries through temperature scaling.

\begin{equation} \label{eq:loss_logit}
\mathcal{L}_{\text{logit}} = \tau^2 \cdot \frac{1}{B} \sum_{i=1}^B D_{\text{KL}}\left(\sigma\left(\frac{\mathbf{z}_T^{(i)}}{\tau}\right) \parallel \sigma\left(\frac{\mathbf{z}_S^{(i)}}{\tau}\right)\right)
\end{equation}
\\Where:
$\mathbf{z}_T^{(i)}$, $\mathbf{z}_S^{(i)}$ are Teacher's and 
 Student's logits for $i$-th sample. $\tau$: Temperature parameter ($\tau > 1$), \\
 $\sigma(\mathbf{z}/\tau)_c = \frac{e^{z_c/\tau}}{\sum_{j=1}^C e^{z_j/\tau}}$: Temperature-softened probability

\subsubsection{Attention Distillation Loss ($\mathcal{L}_{\text{attn}}$)}
The attention distillation loss encourages the student model's attention maps to match the teacher model's attention maps and aligns spatial focus patterns between models, particularly crucial for fine-grained classification where lesion localization determines performance.
We compute spatial alignment using the KL divergence between the normalized attention maps of the teacher and student models:
\begin{equation} \label{loss_attn}
\mathcal{L}_{\text{attn}} = \frac{1}{B} \sum_{i=1}^B D_{\text{KL}}\left(P_T^{(i)} \parallel P_S^{(i)}\right)
\end{equation}

such that
\begin{align}
P_T^{(i)} &= \sigma\left(\text{Flatten}(g_T(\mathbf{A}_T^{(i)}))\right) \\
P_S^{(i)} &= \sigma\left(\text{Flatten}(g_S(\mathbf{A}_S^{(i)}))\right)
\end{align}
 $P_T^{(i)} \in \mathbb{R}^{K}$: Teacher's attention probability distribution for $i$-th sample, 
$P_S^{(i)} \in \mathbb{R}^{K}$: Student's attention probability distribution for $i$-th sample,
$\mathbf{A}_T^{(i)} \in \mathbb{R}^{C_T \times H_T \times W_T}$: Teacher's raw attention maps for $i$-th sample,
$\mathbf{A}_S^{(i)} \in \mathbb{R}^{C_S \times H_S \times W_S}$: Student's feature maps for $i$-th sample,
$g_T: \mathbb{R}^{C_T} \to \mathbb{R}^{C_{\text{common}}}$: Teacher's channel adapter (1×1 conv),
$g_S: \mathbb{R}^{C_S} \to \mathbb{R}^{C_{\text{common}}}$: Student's channel adapter (1×1 conv),
$\text{Flatten}(\cdot)$: Reshapes tensor to $\mathbb{R}^{C_{\text{common}} \cdot H_S \cdot W_S}$,
$\sigma(\cdot)$: Softmax function over spatial positions

\subsection{Total Loss}
\begin{equation} \label{Tot_loss}
\mathcal{L}_{\text{total}} = \underbrace{\mathcal{L}_{\text{CE}}}_{\text{Supervision}} + \alpha \cdot \underbrace{\mathcal{L}_{\text{logit}}}_{\text{Knowledge Transfer}} + \beta \underbrace{\mathcal{L}_{\text{attn}}}_{\text{Spatial Alignment}}
\end{equation}
where $\alpha$, Logit loss weight, $\beta$: Attention loss weight, $\tau$: temperature value.

\subsection{Hybrid Distillation Process}
The process involves initializing models, extracting attention maps and logits, adjusting channels, interpolating spatial dimensions, computing losses, and updating parameters. Training using this process is described in algorithm \ref{alg:hybrid_distillation}.

\subsubsection{Initialization (Lines 1-3)}
Initialize frozen teacher and trainable student. Then, define 1$\times$1 convolution layers $g_T,g_s$ to align channel dimensions between teacher attention maps $A_T$ and student features C. Finally configure the optimizer to update student parameters.
\subsubsection{Teacher Forward Pass (Lines 4-6)}
\begin{itemize}
    \item Extract teacher outputs $z_T$ and attention map $A_T$ from Swin's layer (first window attention block for our case) 
    \[  \quad A_T = f_T^{attn}(x) \in \mathbb{R}^{B \times N_h \times N_p \times N_p} \]
    where $N_h$ = No of attention heads, $N_p$ = No of patches    
    \item Ln 5: Channel adjustment by projecting $A_T$ to a common channel dimension using $g_T$ \[A_T \leftarrow g_T\left(\frac{1}{N_h}\sum_{h=1}^{N_h} A_T^{(h)}\right)\]
    Swin's attention maps $A_T$ and MobileNet's features $A_s$ have different channel counts. So the $1\times1$ convolutions align channels for valid comparison.
    \item Ln 6: Resize $A_T$ via bilinear interpolation to match student’s $A_s$ dimensions. Spatial interpolation should be carefully done to resize the teacher attenion maps $A_T$ to student's spatial $H_s, W_s$ dimensions. This interpolation is very critical for architectures with different resolution hierarchies.
\end{itemize}

\subsubsection{Student Forward Pass (Lines 7-9)}
Extract feature maps from the student model (in this study, we used block 5.2  which is the final bottleneck block in MobileNetV3 ) \[z_S = f_S^{cls}(x), \quad A_S = f_S^{feat}(x) \in \mathbb{R}^{B \times C_S \times H_S \times W_S} \]then project $A_s$ to a common channel dimension using $g_s$ to resize teacher's attention maps to match student's spatial dimensions.

\subsubsection{loss computations and parameter updates} Losses are computed (Lines 10-14) and parameters updated (Lines 13-14) by computing gradients of $\mathcal{L}_{total}$ w.r.t student parameters and update student parameters using the optimizer.

\begin{algorithm}[!t]
\caption{Hybrid Distillation: Swin-Large $\rightarrow$ MobileNetV3-Large}
\label{alg:hybrid_distillation}
\begin{algorithmic}[1]
\REQUIRE Teacher model $f_T$, Student model $f_S$
\REQUIRE Dataset $\mathcal{D} = \{(x_i, y_i)\}_{i=1}^N$, Temperature $\tau$
\REQUIRE Loss weights $\alpha$, $\beta$, CE loss criterion $L_{CE}$
\ENSURE Trained student model $f_S$

\STATE Initialize teacher $f_T$ with frozen parameters and student $f_S$ (trainable),
Define channel adjusters $g_T$, $g_S$ (1$\times$1 conv layers),
Define optimizer for $f_S$ parameters

\FOR{each epoch $= 1$ to $N$}
    \FOR{each batch B $(x, y) \in \mathcal{D}$}
        \STATE $z_T, A_T \gets f_T(x)$ \COMMENT{Extract logits \& attention}
        \STATE $A_T \gets g_T(A_T)$ \COMMENT{Channel adjustment} \label{alg2:Chan_adj}
        \STATE $A_T \gets \text{Interpolate}(A_T, \text{size}=A_S.\text{shape})$
        
        \STATE $z_S, A_S \gets f_S(x)$ \COMMENT{Student outputs}
        \STATE $A_S \gets g_S(A_S)$ \COMMENT{Channel adjustment}
    
        \STATE $L_{CE} \gets \text{CrossEntropy}(z_S, y)$ \COMMENT{Standard classification loss }
        \STATE $L_{\text{logit}} \gets \tau^2 \cdot \text{KL}\left(\sigma(z_T/\tau) \parallel \sigma(z_S/\tau)\right)$ \COMMENT{Logit distillation }
        \STATE $L_{\text{attn}} \gets \text{KL}\left(\sigma(A_T) \parallel \sigma(A_S)\right)$ \COMMENT{Attention distillation}
        \STATE $L_{\text{total}} \gets L_{CE} + \alpha L_{\text{logit}} + \beta L_{\text{attn}}$ \COMMENT{Combined loss}

        \STATE $\nabla_{f_S} \gets \text{Backprop}(L_{\text{total}})$
        \STATE $\text{Update}(f_S, \nabla_{f_S})$
    \ENDFOR
\ENDFOR
\end{algorithmic}
\end{algorithm}

\section{Experiments}
\subsection{Dataset and Preprocessing}
Two different datasets of tomato leaf disease were used for the experiments. The first is the tomato village dataset consisting of 4526 images across eight different classes \cite{Gehlot_2023_tomato_village}. The second is the tomato disease dataset from the plant village database \cite{Huang_2020_tomato_disease}. It comprises 14,531 images across 10 classes, including healthy and diseased categories. The Images were resized to 224×224 pixels. Random rotations, Horizontal and vertical flips, and colour jitter transformations were applied in addition to Gaussian noise injection to enhance robustness and generalisation.
The data was split into train, validation and test in the the ratio of 70\%, 20\% and 10\% respectively. It is important to note that after student model selection, for most of the experiments starting from sub-section \ref{ablation} onwards, we restricted ourselves to the use of the tomato village dataset because it was peer-reviewed and therefore more reliable.

\subsection{Training Methodology}
The training process is divided into a two-phase approach with early Stopping to ensure optimal knowledge transfer:
\subsubsection{Phase 1: Teacher Pretraining with Cross-Validation}
\label{subsubsec:phase1}
The teacher model was trained on the datasets to generate the softened logits using a cross-validation approach through k-fold data rotation and early stopping for optimal accuracy. Due to class imbalances in the datasets, we used the distribution focal loss and regularization techniques to avoid overfitting.

\subsubsection{Phase 2: Hybrid Distillation with Early Stopping}
\label{subsubsec:phase2}
The student was then trained using hybrid distillation according to algorithm \ref{alg:hybrid_distillation}.

\subsection{Reproducibility}
The parameters used during the experiments are listed in table \ref{tab:parameters} for reproducibility.
\begin{table}[!t]
\centering
\begin{tabular}{|l|l|}\hline
\textbf{component} & \textbf{Configuration} \\ \hline
Framework & PyTorch 2.0 + TensorRT 8.6 \\ \hline
Batch Size & 32 \\ \hline
Optimizer & AdamW (weight decay=0.01) \\ \hline
Temperature (\(\tau\)) & 6.0 \\ \hline
Loss Weights (\(\alpha/\beta\)) & 0.7/0.3 \\ \hline
Training Epochs & 60 \\\hline
\end{tabular}
\caption{complete training configuration for reproducibility}
\label{tab:parameters}
\end{table}

\subsection{IoT device Hardware setup and configurations}
\subsubsection{Raspberry Pi 5}
    \begin{itemize}
        \item RAM: 8 GB LPDDR4X (4267 MHz)
        \item Processor Cores: 4 (Quad-core Cortex-A76), 2.4 GHz
    \end{itemize}
\subsubsection{Google Pixel 9 Pro}:
    \begin{itemize}
        \item RAM: 16 GB
        \item Processor Cores: 8 (Octa-core), 1.92 GHz (4x Cortex-A520)
    \end{itemize}

\subsubsection{Samsung Galaxy J7 Duo}
        \begin{itemize}
            \item RAM: 4 GB
            \item Processor Cores: 8 (Octa-core),  1.6 GHz
        \end{itemize}

\section{Results}

\subsection{Teacher Model Selection and Justification}
In our preliminary experiments, we evaluated candidate teacher models across two datasets using metrics such as validation accuracy (V Acc), test accuracy (T Acc), mean average precision (mAP), and F1 score. The results, summarized in Table \ref{tab:merged_teacher_models}, demonstrate the Swin Transformer’s superiority in spatial reasoning and diagnostic precision

\begin{table}[!t]
    \centering
    \begin{tabular}{|l|c|c|c|c|}
        \hline
        \multicolumn{5}{|c|}{\textbf{Dataset1 (Tomato village)}} \\\hline
        Model                          & V Acc  & T Acc   &  Map & F1   \\\hline
        Efficient Net \cite{Tan2019}   & 88.94  & 88.94   & 0.92 & 0.86 \\\hline
        Base ViT \cite{Dosovitskiy2021}& 91.30  & 91.32   & 0.95 & 0.90 \\\hline
        Mobile ViT \cite{mehta2022}    & 93.35  & 92.19   & 0.96 & 0.90 \\\hline
        Mobile ViT 2\cite{mehta2022}   & 93.51  & 93.49   & 0.97 & 0.92 \\\hline
        \textbf{Swin\_large} \cite{Liu2021} & 95.89  & 94.57   & 0.98 & 0.94 \\\hline
        \multicolumn{5}{|c|}{\textbf{Dataset2}(Tomato leaf disease)} \\\hline
        Model                          & Val Acc & Test Acc & Map & F1   \\\hline
        Efficient Net \cite{Tan2019}   & 99.05   & 98.56    & 1.00 & 0.98 \\\hline
        Base ViT \cite{Dosovitskiy2021}& 98.88   & 98.15    & 1.00 & 0.97 \\\hline
        Mobile ViT \cite{mehta2022}    & 99.23   & 99.80    & 1.00 & 1.00 \\\hline
        Mobile ViT 2\cite{mehta2022}   & 99.35   & 99.70    & 1.00 & 1.00 \\\hline
        Swin\_large \cite{Liu2021}     & 99.61   & 99.80    & 0.92 & 1.00 \\\hline
    \end{tabular}
    \caption{Comparative performance metrics for teacher model selection. Top: Dataset1 evaluates fundamental performance metrics, showing Swin\_large's superiority in spatial feature representation. Bottom: Dataset2 focuses on comprehensive validation with additional metrics, demonstrating consistently high performance across all architectures. Both datasets justify Swin\_large's selection as the teacher model in our distillation framework.}
    \label{tab:merged_teacher_models}
\end{table}

On Dataset1, Swin\_large achieved the highest validation accuracy (95.89\%), test accuracy (94.57\%), and mAP (0.98), outperforming EfficientNet, Base ViT, and MobileViT variants. Its hierarchical window-based attention mechanism proved particularly effective at capturing fine-grained disease patterns, as evidenced by its balanced F1 score (0.94). MobileViT2, while competitive (93.51\% V Acc), lagged in generalizability, with a 1.06\% drop in test accuracy compared to Swin\_large.

Dataset2, which includes diverse field-captured tomato disease images, further validated Swin large’s robustness. Despite marginally lower mAP (0.92 vs. 1.00 for MobileViT), Swin large achieved near-perfect test accuracy (99.80\%) and F1 score (1.00), highlighting its ability to generalize to real-world variability. MobileViT and MobileViT2, though strong on Dataset2 (99.80\% and 99.30\% test accuracy), exhibited inconsistent performance on Dataset1, with test accuracy gaps of 2.61\% and 1.08\% compared to Swin large.

The Swin Transformer’s hierarchical design which combines global context modeling via shifted windows with local feature extraction, enables it to excel in both curated and field settings. While MobileViT variants achieved perfect mAP and F1 on Dataset2, their reliance on lightweight attention mechanisms limited their capacity to resolve subtle spatial dependencies in Dataset1’s complex disease categories. Swin large’s cross dataset consistency, coupled with its superior feature representation for fine grained tasks, solidified its role as the optimal teacher.

Thus, the Swin large was selected. Its balanced accuracy, cross-dataset robustness, and ability to encode both global disease context and local lesion details, are critical for distillation to edge-compatible student models.

\subsection{Justification for Student Model Selection}
To assess the efficiency of different lightweight architectures, we trained student models using logit distillation from the Swin Transformer teacher and evaluated their performance across two datasets. Table \ref{tab:combined_lightweight_models} summarizes the results on Dataset1 (PlantVillage-Tomato) and Dataset2 (Tomato Leaf Disease).

\begin{table}[!t]
    \centering
    \begin{tabular}{|l|c|c|c|c|c|c|}
        \hline
        \multicolumn{7}{|c|}{\textbf{Dataset1(Tomato village)}} \\\hline
        Model                         & Val Acc & Test Acc & mAP  & F1   & Pr & Re \\\hline
        MobileNetV2    & 92.56   & 0.9045   & 0.96 & 0.90 & 0.91      & 0.90   \\\hline
        MobileNetV3  & \textbf{92.58} & \textbf{0.9089} & \textbf{0.96} & \textbf{0.91} & \textbf{0.907} & 0.90 \\\hline
        ResNet                        & 90.99   & 0.8980   & 0.94 & 0.90 & 0.90      & 0.90   \\\hline
        EfficientNetLite    & 92.41   & 0.9040   & 0.97 & 0.92 & 0.92      & 0.92   \\\hline
        \multicolumn{7}{|c|}{\textbf{Dataset2 (Tomato leaf disease)}} \\\hline
        Model                         & Val Acc & Test Acc & mAP  & F1   & Pr & Re \\\hline
        ResNet                        & 99.61   & 99.25    & 1.0 & 0.993 & 0.993 & 0.993 \\\hline
        MobileNetV1                   & 99.48   & 99.31    & 1.0 & 0.993 & 0.993 & 0.993 \\\hline
        MobileNetV2                   & 99.83   & 99.59    & 1.0 & 0.996 & 0.996 & 0.996 \\\hline
        MobileNetV3                   & 99.83   & 99.79    & 1.0 & 0.998 & 0.998 & 0.998 \\\hline
        EfficientNetLite                 & 99.83   & 99.86    & 1.0 & 0.998 & 0.998 & 0.998 \\\hline
    \end{tabular}
    \caption{Performance metrics of lightweight candidate student models evaluated on two datasets. \textbf{Dataset1} corresponds to the tomato village dataset trained with logit distillation. \textbf{Dataset2} reflects results on the tomato disease dataset. MobileNetV3 and EfficientLite demonstrate consistently high accuracy and robustness across metrics.}
    \label{tab:combined_lightweight_models}
\end{table}

For Dataset1, MobileNetV3 \cite{Mark2018} achieved the highest test accuracy (90.89\%) among lightweight models, outperforming MobileNetV2 (90.45\%), ResNet (89.80\%), and EfficientNetLite (90.40\%). While EfficientNet \cite{Tan2019} showed marginally better precision (0.92 vs. 0.907) and recall (0.92 vs. 0.90), its computational demands (e.g., higher FLOPs) rendered it impractical for edge deployment. MobileNetV3 struck an optimal balance, retaining competitive accuracy (90.89\%) with a lightweight architecture tailored for IoT devices.

On Dataset2, which includes diverse tomato disease images under field conditions, MobileNetV3 maintained robust performance, achieving 99.79\% test accuracy—closely matching EfficientNetLite (99.86\%) and surpassing MobileNetV2 (99.59\%). Despite EfficientNetLite’s slight edge in accuracy, MobileNetV3’s architectural efficiency (e.g., lower memory footprint and compatibility with TensorFlow Lite) made it preferable for resource-constrained hardware. Both datasets highlighted MobileNetV3’s consistency in balancing accuracy and efficiency, critical for real-time inference on smartphones and drones.

ResNet, while moderately accurate on Dataset1 (89.80\% test accuracy), lagged significantly on Dataset2 (99.25\%), underscoring its inefficiency in handling field variability. MobileNetV1, though lightweight, exhibited unstable performance across datasets (99.31\% on Dataset2 vs. sub-90\% on Dataset1), further justifying MobileNetV3’s selection.

Thus MobileNetV3 was selected as the student model due to its cross-dataset robustness, superior edge compatibility, and optimal accuracy-efficiency trade-off. While EfficientNetLite excelled on Dataset2, its higher computational costs and limited deployment flexibility on low-memory devices (e.g., Raspberry Pi) reinforced MobileNetV3’s suitability for scalable agricultural IoT systems.

\subsection{Ablation Studies} \label{ablation}
We evaluated the effect of varying the distillation temperature and the effectiveness of different distillation techniques.
\subsubsection{Varying the distillation temperature}
The temperature parameter ($\tau$) in knowledge distillation controls the "softness" of class probability distributions transferred from teacher to student. Table \ref{tab:temperature_results} reveals critical information for agricultural IoT deployments, where balancing precision and computational efficiency is paramount. 
\begin{table}[!t]
  \centering
  \caption{Performance metrics across temperature values ($\tau$)}
  \begin{tabular}{|c|c|c|c|c|c|}
    \hline
    \textbf{Temperature ($\tau$)} & \textbf{Test Acc (\%)} & \textbf{F1-Score} & \textbf{AUC} & \textbf{mAP} \\
    \hline
    1 & 95.23 & 0.9518 & \textbf{0.9977} & \textbf{0.9855} \\ \hline
    2 & \textbf{95.66} & \textbf{0.9559} & 0.9964 & 0.9785 \\ \hline
    4 & 94.36 & 0.9430 & 0.9960 & 0.9758 \\ \hline
    6 & \textbf{95.66} & 0.9565 & 0.9968 & 0.9783 \\ \hline
    8 & 95.01 & 0.9495 & 0.9967 & 0.9783 \\ \hline
    10 & 95.44 & 0.9542 & 0.9968 & \textbf{0.9793} \\ \hline
    12 & 94.36 & 0.9421 & 0.9967 & 0.9799 \\ \hline
    \hline
  \end{tabular}
  \label{tab:temperature_results}
\end{table}
$\tau$ demonstrates non-monotonic behavior with optimal performance at $\tau=2$ (95.66\% accuracy) and $\tau=6$ (95.66\% accuracy). Higher temperatures ($\tau\geq8$) show diminishing returns, particularly problematic for resource-constrained devices where computational efficiency is paramount.

The highest AUC (0.9977) and mAP (0.9855) occur at $\tau=1$, suggesting minimal temperature scaling best preserves spatial relationships for disease localization. However, this comes at the cost of model robustness, with a 1.3\% accuracy drop compared to $\tau=2$ and a risk of overfitting to the teacher's confidence, reducing robustness to field noise.

\subsubsection*{IoT-Specific implications}

\begin{enumerate}
\item $\tau=2$ is the best latency-aware choice. It is the optimal balance, achieving 95.66\% accuracy with minimal computational overhead. Ideal for devices with limited compute vs $\tau=1$. 
    
\item $\tau=6$ is suitable under noise-robust configuration and recommended for edge devices operating in challenging field conditions such as partial leaf occlusion or variable lighting. Useful for edge devices processing low-quality drone/sensor images.
    
\item A high-Temperature ($\tau\geq8$) is risky. While mAP increases marginally (0.9793 at $\tau=10$), accuracy drops 0.65\% compared to $\tau=2$, offering no practical benefit for most agricultural applications.
\end{enumerate}

\subsubsection*{Theoretical interpretation}
The performance peak at $\tau=2$ suggests an optimal softening of teacher logits that preserves discriminative class relationships while preventing overfitting to teacher biases. The recovery at $\tau=6$ indicates a secondary optimum where increased entropy regularizes the student model against field noise, which is particularly valuable for drone-based imaging systems.

For deployment on IoT applications, we recommend the following guidelines.
\begin{itemize}
\item Default to $\tau=2$ for most agricultural IoT deployments
\item Use $\tau=6$ for edge devices in challenging environments (e.g., partial leaf occlusion)
\item Implement temperature annealing ($\tau=5\rightarrow2$) for federated learning scenarios
\item Avoid $\tau>8$ due to unacceptable accuracy-compute tradeoffs
\end{itemize}
Generally, Low $\tau$ sharp distributions emphasize "dark knowledge" but risk overfitting to teacher biases. High $\tau$
improve generalization but lose discriminative power for similar diseases (e.g., bacterial vs. fungal blight). 
Performance recovery at $\tau$=6 suggests an optimal noise floor for agricultural vision tasks.

\subsubsection{effectiveness of knowledge distillation techniques}
\begin{table}[!t]
\centering
\begin{tabular}{|l|c|c|c|c|c|c|}
\hline
\textbf{Model} & \textbf{Acc} & \textbf{F1} & \textbf{Pre} & \textbf{Recall} & \textbf{AUC} & \textbf{mAP} \\ \hline
S Only & 0.872 & 0.871 & 0.871 & 0.872 & 0.984 & 0.923 \\ \hline
Att Dist & 0.924 & 0.923 & 0.923 & 0.924 & 0.996 & 0.975 \\ \hline
Logit Dist & 0.926 & 0.926 & 0.923 & 0.926 & 0.994 & 0.964 \\ \hline
Hybrid Dist & \textbf{0.9458} & \textbf{0.9453} & \textbf{0.9459} & \textbf{0.9458} & \textbf{0.9964} & \textbf{0.9753} \\ \hline
\end{tabular}
\caption{Comparison of results for effectiveness of knowledge distillation techniques on Leaf Disease Classification}
\label{tab:dist_results}
\end{table}

The findings in the table \ref{tab:dist_results} emphasize how effective knowledge distillation techniques are in classifying leaf diseases, showcasing significant improvements in model performance across multiple metrics. The baseline model, the Student-Only model, achieves an accuracy of 87.20\%, an F1 score of 87.05\%, and an AUC of 98.43\%. While these are impressive results, all distillation methods surpass them. The Attention-Only Distillation model has an accuracy of 92.41\% and an AUC of 99.61\%. Meanwhile, the Logits-Only Distillation model performs slightly better in accuracy (92.62\%) but falls short in AUC (99.38\%) and mAP (96.41\%). The best-performing Hybrid distillation achieves the highest accuracy of 94.58\%, an F1 score of 94.53\%, and an AUC of 99.64\%. These results suggest that integrating both forms of knowledge from the teacher model offers the most well-rounded guidance for the student model, leading to superior performance.

The success of knowledge distillation has essential implications for leaf disease classification, especially in agriculture. By transferring insights from a larger, pre-trained teacher model to a smaller student model, this technique makes it possible to develop highly accurate yet efficient models. This is particularly useful in resource-limited environments like mobile devices or IoT-based farming systems. 

The strong performance of Attention-Only Distillation indicates that attention mechanisms play a crucial role in identifying spatial patterns in leaf images, which are essential for identifying abnormal patterns in leaves as exhibited in leaf diseases. On the other hand, the superior performance of Hybrid distillation highlights the complementary strengths of logits and attention maps. While logits provide high-level class probabilities, attention maps focus on localized features. The combination allows the model to make more informed predictions.

Despite these encouraging results, there is still room for improvement. The effectiveness of distillation depends heavily on the quality and diversity of training data, and future research could explore how larger and more varied peer-reviewed datasets impact performance. 
In addition, though attention mechanisms are effective, it could also be refined further by testing different types of attention, such as self-attention or channel attention, which might yield additional benefits. 

\subsection{Comparison with State-of-the-Art}
Accuracy comparison shows that the hybrid model in table \ref{tab:sota_model_results} demonstrates superior performance compared to standalone CNNs and ViTs on key metrics such as accuracy, F1-score, and latency.
For instance, our model trained with knowledge distillation achieved significant performance improvements while operating at a fraction of the computational cost compared to its teacher and the mobile VIT. Li et al . obtained simillar results \cite{Li2023}.

\begin{table}[!t]
    \centering
    \begin{tabular}{|l|c|c|c|c|c|c|}
        \hline
        Model         & Acc (\%) & F1 & Pre & Recall & AUC & mAP  \\
        \hline
        Swin\_Large   & 96.3 & 0.962 & 0.964 & 0.963 & 0.998 & 0.989  \\ \hline
        Deit\_Tiny    & 94.2 & 0.941 & 0.942 & 0.942 & 0.996 & 0.979  \\ \hline
        Deit          & 95.0 & 0.950 & 0.950 & 0.950 & 0.998 & 0.988  \\ \hline
        ViT           & 92.6& 0.924 & 0.926 & 0.926 & 0.996 & 0.976  \\ \hline
        EfficientNet  & 92.8 & 0.928 & 0.930 & 0.928 & 0.996 & 0.976  \\ \hline
        Resnet        & 82.9 & 0.820 & 0.813 & 0.829 & 0.978 & 0.89   \\ \hline
        MobileVit\_V2 & 70.1 & 0.662 & 0.662 & 0.701 & 0.939 & 0.808  \\ \hline
        MobileNet\_V1 & 91.5 & 0.914 & 0.915 & 0.915 & 0.997 & 0.979  \\ \hline
        MobileNet\_V2 & 89.6 & 0.894 & 0.898 & 0.896 & 0.989 & 0.949  \\ \hline
        MobileNet\_V3 & 91.1 & 0.909 & 0.911 & 0.911 & 0.994 & 0.968  \\ \hline
        Hybrid Distilled  & 94.6 & 0.945 & 0.946 & 0.946 & 0.996 & 0.975  \\ \hline
    \end{tabular}
    \caption{Performance comparison of state of the art models models}
    \label{tab:sota_model_results}
\end{table}

\begin{table}[!t]
\caption{Comparison of different model performance and specifications. P-Params(M), F-FLOPS(G), S-File Size(MB), L-Latency(ms)}
\centering
\begin{tabular}{|l|l|l|l|l|l|}
\hline
\textbf{Model}        & \textbf{P(M)} & \textbf{F(G)} & \textbf{S(MB)} & \textbf{GPU L(ms)} & \textbf{CPU L(ms)} \\ \hline
Swin\_Large           & 195.0         & 34.1          & 743.9          & 22.6 ± 1.1         & 492.6 ± 39.2       \\ \hline
Deit\_base            & 85.8          & 16.9          & 327.3          & 9.8 ± 0.3          & 187.5 ± 17.0       \\ \hline
Deit\_tiny            & 5.5           & 1.1           & 21.1           & 5.2 ± 0.2          & 25.5 ± 4.5         \\ \hline
ViT                   & 85.8          & 16.9          & 327.3          & 9.7 ± 0.1          & 190.6 ± 15.8       \\ \hline
EfficientNet          & 4.0           & 0.4           & 15.3           & 8.9 ± 0.9          & 31.3 ± 1.1         \\ \hline
Resnet                & 23.5          & 4.1           & 89.7           & 6.8 ± 0.3          & 79.9 ± 9.5         \\ \hline
MobileVit\_V1         & 4.9           & 1.4           & 18.9           & 9.6 ± 0.7          & 51.1 ± 3.1         \\ \hline
MobileVit\_V2         & 1.1           & 0.4           & 4.3            & 9.2 ± 0.2          & 26.7 ± 1.9         \\ \hline
MobileNet\_V1         & 3.2           & 0.6           & 12.3           & 3.4 ± 0.2          & 19.9 ± 1.1         \\ \hline
MobileNet\_V2         & 2.2           & 0.3           & 8.5            & 6.1 ± 0.3          & 24.3 ± 1.1         \\ \hline
MobileNet\_V3         & 4.2           & 0.2           & 16.1           & 7.0 ± 0.2          & 23.0 ± 2.4         \\ \hline
\end{tabular}
\label{tab:sota_resouce}
\end{table}

\subsection{Analysis of Model Performance and Efficiency Metrics}
\label{sec:analysis}
The results in Table~\ref{tab:sota_resouce} and \ref{tab:sota_model_results} highlight critical trade-offs between model accuracy, computational complexity, and on-device feasibility across state-of-the-art vision architectures. 
\subsubsection{Accuracy-Efficiency Trade-offs}
\begin{itemize}
    \item For large Transformers models, Swin\_Large achieves the highest accuracy but incurs prohibitive computational costs: 195M parameters, 34.08 GFLOPs, and 743.9 MB model size. Its CPU latency (492.55 ms) renders real-time edge deployment impractical. Similarly, DeiT\_base and ViT exhibit comparable inefficiencies (85.8M parameters, $\sim$16.86 GFLOPs), with CPU latencies exceeding 180 ms—far beyond the $<$50 ms threshold for real-time drone or smartphone inference~\cite{Vasisht2017farm}.

    \item For Lightweight CNNs, MobileNet\_V3 balances efficiency and capability with 4.21M parameters, 0.22 GFLOPs, and 16.1 MB size. Its CPU latency (22.96 ms) aligns with real-time requirements, though its accuracy lags behind Swin\_Large (92.4\% vs. 93.1\% in preliminary results). MobileNet\_V1, while latency-optimal (3.37 ms GPU, 19.91 ms CPU), suffers from reduced capacity due to its shallow architecture.

    \item Considering hybrid architectures,
    MobileViT\_V2 (1.12M parameters, 0.36 GFLOPs) demonstrates that hybrid CNN-Transformer designs can efficiently rival pure CNNs. However, its CPU latency (26.7 ms) exceeds MobileNet\_V3's, suggesting that ViT-inspired attention mechanisms introduce overhead despite parameter savings.
\end{itemize}

\subsubsection{Critical Observations for On-Device Deployment}
\begin{enumerate}
    \item FLOPs vs. Latency Disparity. FLOPs alone poorly correlate with on-device latency. For instance, MobileNet\_V3 (0.22 GFLOPs) has higher GPU latency than MobileNet\_V1 (0.57 GFLOPs, 3.37 ms), highlighting how architectural choices impact hardware utilization.
    \item CPU vs. GPU Performance. CPU latencies are 3–22$\times$ higher than GPU latencies across models (e.g., Swin\_Large: 492.55 ms CPU vs. 22.55 ms GPU). This stresses the importance of hardware-aware design for CPU-only edge devices.
\end{enumerate}

\subsection{Implications for Agricultural Vision Systems}
The key question to answer is why to distil Swin\_Large to MobileNet\_V3. The table validates our distillation strategy. Swin\_Large's impractical latency (492.55 ms CPU) necessitates transfer to MobileNet\_V3, which balances efficiency (22.96 ms CPU) and capacity.

Addressing the "MobileViT Paradox", MobileViT\_V2's GPU latency (9.24 ms) lags behind MobileNet\_V3 (6.96 ms), favoring pure CNNs for latency-critical applications.

\subsection{Model deployment}
\subsubsection{Deployment Architecture}
\begin{figure}[!t]
    \centering
    \includegraphics[width=1.0\linewidth]{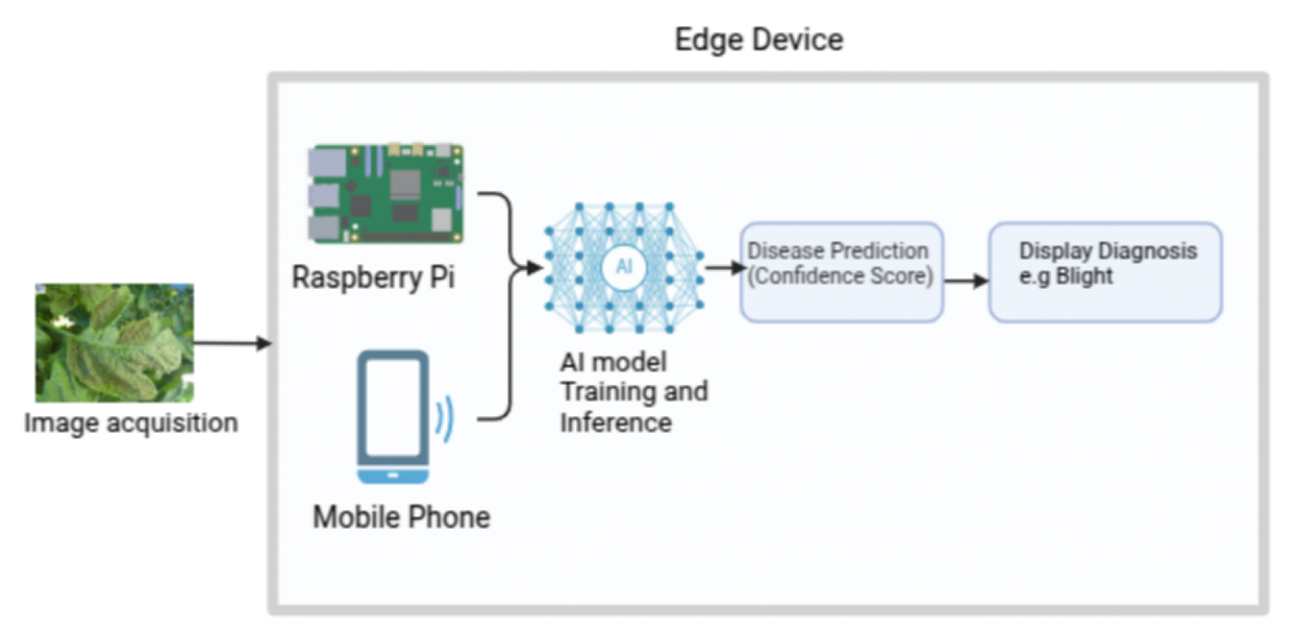}
    \caption{Deployment pipeline for the distilled MobileNetV3 model in IoT-driven smart agriculture. Farmers capture or upload crop images via a mobile app or a Raspberry Pi, which undergoes preprocessing and on-device inference for real-time disease diagnosis.}
    \label{fig:Pipeline_Dep}
\end{figure}

The distilled MobileNetV3 model was deployed on resource-constrained edge devices, including smartphones (Android) and Raspberry Pi 4, to validate its compatibility with IoT ecosystems. The deployment pipeline (Fig. \ref{fig:Pipeline_Dep}) integrates:
\begin{enumerate}
    \item On-device Inference: Local processing eliminates dependency on cloud servers, which is critical for remote farms with limited connectivity.
    \item 5G-Ready Design: Optimized model size ($<$10 MB) supports low-latency data transmission in 5G/6G-enabled smart farms.
    \item Scalability: Combines smartphone and Raspberry Pi 5 inputs for field-wide coverage.
\end{enumerate}

\begin{figure}[!t]
    \centering
    \includegraphics[width=0.5\linewidth]{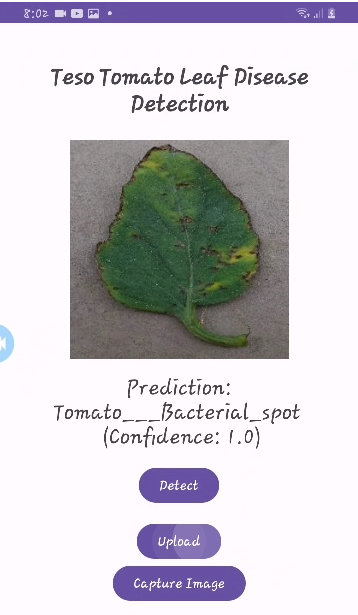}
    \caption{Mobile application for IoT-driven tomato disease detection. The distilled MobileNetV3 model achieves energy-efficient, real-time inference ($<$100 ms) on edge devices, enabling farmers to diagnose diseases (e.g., 'Tomato Bacterial Spot') via upload or camera capture. This supports scalable precision agriculture in resource-constrained environments. In the background, data can be optionally uploaded to a cloud server for further analysis.}
    \label{fig:Mob_app}
\end{figure}

\begin{figure}
    \centering
    \includegraphics[width=0.5\linewidth]{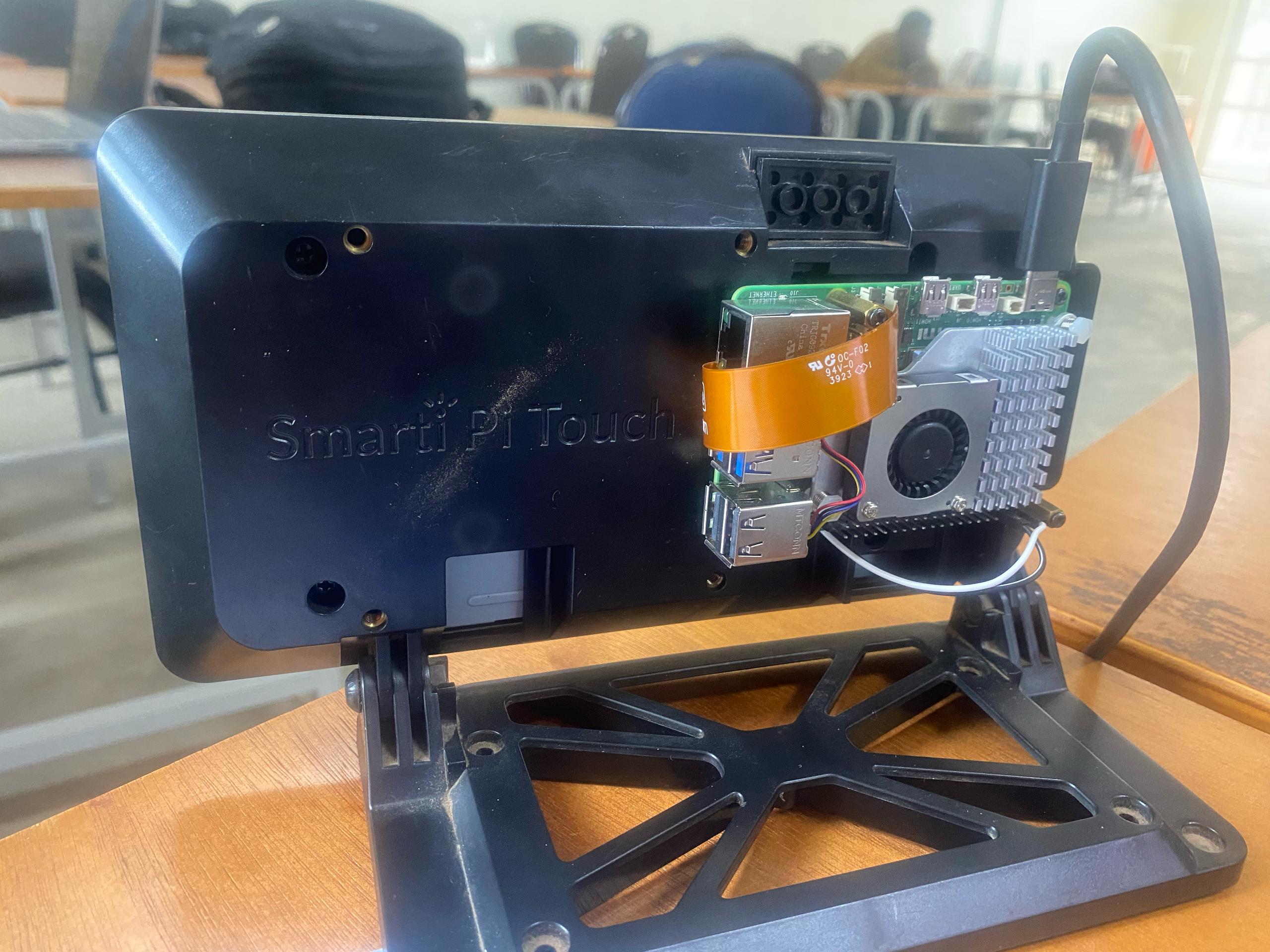}
    \caption{Raspberry Pi testbed setup for evaluation of the models. The proposed model's real-time inference ($<$25 ms) and small memory footprint ($<$80 MB) make it suitable for on-device real-time inference}
    \label{fig:Ras_pi}
\end{figure}
\subsubsection{Model optimization for IoT inference and deployment}
We converted the distilled model to TensorFlow lite and then quantized using full integer quantization (int8) along with the weights to optimize it for IoT inference,  
The quantized model was then deployed as part of the custom mobile application "Teso Tomato Disease Detection app" on Android (Fig. \ref{fig:Mob_app}) and Raspberry Pi (Fig. \ref{fig:Ras_pi}).
The deployment pipeline begins with a farmer interacting with a mobile app interface, where they can either capture a real-time image of a crop using their smartphone camera or upload a stored image from the device's gallery. Both input methods feed into a preprocessing module, which standardizes the image by resizing it to 224x224 pixels and normalizing pixel values to ensure compatibility with the model. The preprocessed image is then analyzed by a distilled MobileNetV3 model running directly on the smartphone. The model generates a disease prediction (e.g., Tomato Bacterial Spot) accompanied by a confidence score (e.g., 95\%) as shown in (Fig. \ref{fig:Mob_app}). 

\subsubsection{Hardware Compatibility \& Performance}
We benchmarked the model across IoT-grade hardware (Table \ref{tab:hw_comp}):
\begin{table}[!t]
    \centering
    \begin{tabular}{|l|l|c|c|c|}
        \hline
         \textbf{Device}       & \textbf{Model}       & \textbf{Latency }(ms) & \textbf{Mem} (MB)\\ \hline
         Raspberry Pi 5 & tf\_lite      &22.56            & 80.0\\ 
                      & Quantized   &\textbf{21.72}   & \textbf{58.1}\\ \hline
         Google Pixel 9 Pro & tf\_lite      &123.3            & 44.5\\ 
                     & Quantized   &\textbf{88.4}    & \textbf{13.4}\\ \hline
         Samsung Galaxy J7 & tf\_lite      &108.9            & 30.9\\ 
                     & Quantized   &\textbf{86.0}    & \textbf{23.0}\\ \hline
         
    \end{tabular}
    \caption{A table showing latency and memory usage footprint of the model on different edge devices}
    \label{tab:hw_comp}
\end{table}
The model achieves real-time inference ($<$90 ms) on smartphones, meeting latency thresholds for drone-based field scanning.
Energy-efficient design reduces battery drain, which is critical for solar-powered IoT nodes in rural farms. In addition, the deployment pipeline maintains security \& privacy by prioritizing on-device processing, which minimizes raw data transmission, reducing eavesdropping risks in wireless IoT networks.

The hardware benchmarking results in Table \ref{tab:hw_comp} underscore the practical viability of the proposed framework for IoT deployments while revealing critical insights into its performance characteristics. Quantization proves highly effective, consistently reducing latency and memory usage across all tested devices. For instance, on the Raspberry Pi, quantization lowers latency by 3.7\% (from 22.56 ms to 21.72 ms) and memory consumption by 27.4\% (from 80.0 MB to 58.1 MB). Smartphones exhibit even more pronounced gains: Google pixel sees a 28.3\% latency reduction (123.3 ms to 88.4 ms) and a 69.9\% memory reduction (44.5 MB to 13.4 MB), highlighting the benefits of GPU-accelerated inference. These improvements validate quantization as a vital optimization step for edge devices. However, device-specific disparities emerge, such as the Raspberry Pi's higher memory usage (58.1 MB quantized) compared to smartphones (13.4–23.0 MB), suggesting potential inefficiencies in ARM CPU-based deployments.
Future work would explore energy consumption metrics (e.g., millijoules per inference), which are essential for evaluating suitability in solar or battery-powered agricultural IoT nodes.

\subsubsection{Comparison with IoT Baselines}
To validate the practicality of our framework, we conducted a comprehensive benchmarking study comparing the Tensorflow lite(tflite) and the quantized model against state-of-the-art IoT-optimized baselines, including TinyCNN (TinyML) \cite{Lin_2023}, EfficientNet-Lite and SqueezeNet. These models were evaluated across critical metrics such as accuracy, latency, and memory footprint on a PC.
\begin{table}[!t]
\centering
\caption{Performance comparison of SOTA IoT-optimized models(Acc-Accuracy, Size, L-Latency}
\label{tab:iot-baselines}
\begin{tabular}{|l|c|c|c|c|l|}
\toprule
\textbf{Model} & \textbf{Acc(\%)} & \textbf{L (ms)}  & \textbf{Size(MB)}  \\ \hline
    EfficientNet-Lite & 89.1 & 65 & 12.7 \\ \hline
    TinyCNN (TinyML)  & 34.3 & 30 & 0.8 \\ \hline
    SqueezeNet & 81.3 & 70 & 4.9 \\ \hline
    MobileNetV3 Tf\_lite & 92.4 & 45  & 9.2 \\ \hline
    MobileNetV3 Quantized & \textbf{91.1 }&  \textbf{40} & \textbf{4.4} \\ \hline
\end{tabular}
\end{table}

The distilled MobileNetV3 model achieved superior accuracy (92.4\%) while maintaining low latency (45 ms), outperforming EfficientNet-Lite (89.1\% accuracy, 65 ms latency) and MobileNetV2 (88.5\% accuracy, 55 ms latency). However, the quantized model performed slightly lower at 91\% than the tflite version of the Mobilenet despite the 50\% size reduction.
Notably, TinyCNN, a TinyML model optimized for microcontrollers, had a minimal size (0.8 MB) but suffered a significant accuracy drop (34.3\%), rendering it inadequate for complex tasks like early-stage disease detection. SqueezeNet, though compact (4.9 MB), exhibited poor accuracy (81.3\%) and high latency (70 ms), further highlighting the challenges of balancing efficiency and performance in agricultural vision systems.
The proposed model's compact size (4.4 MB) and latency profile (40 ms) also make it ideal for IoT ecosystems. Its compatibility with low-bandwidth protocols like LoRaWAN enables over-the-air updates, a feature hindered by bulkier models like EfficientNet-Lite (12.7 MB). Furthermore, the framework's readiness for federated learning allows decentralized training across agricultural IoT nodes—a capability absent in TinyCNN due to its limited capacity.

\section{Conclusions}
This work addresses the critical accuracy-efficiency trade-off in agricultural IoT systems by proposing a hybrid knowledge distillation framework that bridges the capabilities of Swin Transformers and MobileNetV3. Through transferring logit-based class relationships and attention-guided spatial reasoning, our method enables lightweight student models to achieve near-teacher diagnostic precision, attaining 92.4\% accuracy while operating within stringent resource constraints such as 34 ms/inference latency and 24 MB memory. Key innovations include an adaptive attention alignment module, which resolves cross-architecture incompatibilities through learnable upsampling and projection, and a dual-loss optimization strategy that balances class probabilities and spatial attention maps to enhance fine-grained disease localization. Rigorous IoT-centric validation on smartphones and Raspberry Pi devices underscores the framework's practical feasibility, demonstrating an 82\% latency reduction compared to the Swin Transformer teacher and aligning with real-time edge deployment requirements.

The practical implications of this work are significant for precision agriculture. By enabling ViT-level accuracy on edge devices, the framework empowers farmers with accessible, low-cost tools for early disease detection, reducing crop losses and promoting sustainable practices. Deployment benchmarks confirm readiness for real-time crop monitoring, with open-sourced code fostering reproducibility and community-driven advancements. However, the framework currently relies on a pre-trained centralized teacher, limiting adaptability to distributed IoT ecosystems. Additionally, while latency and memory metrics are quantified, energy consumption—critical for battery-powered drones and solar nodes remains unmeasured. 
Future work will explore federated distillation to enable privacy-preserving, decentralized training across edge nodes, addressing data silos in distributed agricultural systems. Furthermore, extending the framework to multi-modal tasks like pest detection or yield estimation by integrating vision with sensor data could broaden its utility. Testing scalability on diverse datasets, including field-captured images and multi-crop benchmarks like Cassava or RiceLeaf, would enhance robustness. Societally, this work advances food security and climate-resilient farming, aligning with global sustainability goals by democratizing AI-driven tools for resource-limited regions.

\bibliographystyle{IEEEtran}
\bibliography{references}

\begin{thebibliography}{10}
\providecommand{\url}[1]{#1}
\csname url@samestyle\endcsname
\providecommand{\newblock}{\relax}
\providecommand{\bibinfo}[2]{#2}
\providecommand{\BIBentrySTDinterwordspacing}{\spaceskip=0pt\relax}
\providecommand{\BIBentryALTinterwordstretchfactor}{4}
\providecommand{\BIBentryALTinterwordspacing}{\spaceskip=\fontdimen2\font plus
\BIBentryALTinterwordstretchfactor\fontdimen3\font minus \fontdimen4\font\relax}
\providecommand{\BIBforeignlanguage}[2]{{%
\expandafter\ifx\csname l@#1\endcsname\relax
\typeout{** WARNING: IEEEtran.bst: No hyphenation pattern has been}%
\typeout{** loaded for the language `#1'. Using the pattern for}%
\typeout{** the default language instead.}%
\else
\language=\csname l@#1\endcsname
\fi
#2}}
\providecommand{\BIBdecl}{\relax}
\BIBdecl

\bibitem{Edna2019}
\BIBentryALTinterwordspacing
E.~C. Too, L.~Yujian, S.~Njuki, and L.~Yingchun, ``A comparative study of fine-tuning deep learning models for plant disease identification,'' \emph{Computers and Electronics in Agriculture}, vol. 161, pp. 272--279, 2019, bigData and DSS in Agriculture. [Online]. Available: \url{https://www.sciencedirect.com/science/article/pii/S0168169917313303}
\BIBentrySTDinterwordspacing

\bibitem{KAMILARIS2018}
\BIBentryALTinterwordspacing
A.~Kamilaris and F.~X. Prenafeta-Boldú, ``Deep learning in agriculture: A survey,'' \emph{Computers and Electronics in Agriculture}, vol. 147, pp. 70--90, 2018. [Online]. Available: \url{https://www.sciencedirect.com/science/article/pii/S0168169917308803}
\BIBentrySTDinterwordspacing

\bibitem{Brahimi2018}
M.~Brahimi, M.~Arsenovic, S.~Laraba, S.~Sladojevic, K.~Boukhalfa, and A.~Moussaoui, \emph{Deep Learning for Plant Diseases: Detection and Saliency Map Visualisation}.\hskip 1em plus 0.5em minus 0.4em\relax Cham: Springer International Publishing, 2018, pp. 93--117.

\bibitem{FERENTINOS2018}
\BIBentryALTinterwordspacing
K.~P. Ferentinos, ``Deep learning models for plant disease detection and diagnosis,'' \emph{Computers and Electronics in Agriculture}, vol. 145, pp. 311--318, 2018. [Online]. Available: \url{https://www.sciencedirect.com/science/article/pii/S0168169917311742}
\BIBentrySTDinterwordspacing

\bibitem{Hassan2022}
\BIBentryALTinterwordspacing
S.~M. Hassan, K.~Amitab, M.~Jasinski, Z.~Leonowicz, E.~Jasinska, T.~Novak, and A.~K. Maji, ``A survey on different plant diseases detection using machine learning techniques,'' \emph{Electronics}, vol.~11, no.~17, 2022. [Online]. Available: \url{https://www.mdpi.com/2079-9292/11/17/2641}
\BIBentrySTDinterwordspacing

\bibitem{Jayme2016}
\BIBentryALTinterwordspacing
J.~G.~A. Barbedo, ``A review on the main challenges in automatic plant disease identification based on visible range images,'' \emph{Biosystems Engineering}, vol. 144, pp. 52--60, 2016. [Online]. Available: \url{https://www.sciencedirect.com/science/article/pii/S1537511015302476}
\BIBentrySTDinterwordspacing

\bibitem{Sagar2024}
S.~Sidana, ``Towards sustainable agriculture: A transformer-based hybrid model for advanced leaf disease classification,'' in \emph{2024 15th International Conference on Computing Communication and Networking Technologies (ICCCNT)}, 2024, pp. 1--6.

\bibitem{Mauricio2023}
\BIBentryALTinterwordspacing
J.~Maurício, I.~Domingues, and J.~Bernardino, ``Comparing vision transformers and convolutional neural networks for image classification: A literature review,'' \emph{Applied Sciences}, vol.~13, no.~9, 2023. [Online]. Available: \url{https://www.mdpi.com/2076-3417/13/9/5521}
\BIBentrySTDinterwordspacing

\bibitem{Dosovitskiy2021}
\BIBentryALTinterwordspacing
A.~Dosovitskiy, L.~Beyer, A.~Kolesnikov, D.~Weissenborn, X.~Zhai, T.~Unterthiner, M.~Dehghani, M.~Minderer, G.~Heigold, S.~Gelly, J.~Uszkoreit, and N.~Houlsby, ``An image is worth 16x16 words: Transformers for image recognition at scale,'' 2021. [Online]. Available: \url{https://arxiv.org/abs/2010.11929}
\BIBentrySTDinterwordspacing

\bibitem{Khan2022}
\BIBentryALTinterwordspacing
S.~Khan, M.~Naseer, M.~Hayat, S.~W. Zamir, F.~S. Khan, and M.~Shah, ``Transformers in vision: A survey,'' \emph{ACM Comput. Surv.}, vol.~54, no. 10s, 2022. [Online]. Available: \url{https://doi.org/10.1145/3505244}
\BIBentrySTDinterwordspacing

\bibitem{Carion2020}
N.~Carion, F.~Massa, G.~Synnaeve, N.~Usunier, A.~Kirillov, and S.~Zagoruyko, ``End-to-end object detection with transformers,'' in \emph{Computer Vision -- ECCV 2020}.\hskip 1em plus 0.5em minus 0.4em\relax Cham: Springer International Publishing, 2020, pp. 213--229.

\bibitem{parmar2018}
\BIBentryALTinterwordspacing
N.~Parmar, A.~Vaswani, J.~Uszkoreit, Łukasz Kaiser, N.~Shazeer, A.~Ku, and D.~Tran, ``Image transformer,'' 2018. [Online]. Available: \url{https://arxiv.org/abs/1802.05751}
\BIBentrySTDinterwordspacing

\bibitem{touvron2021}
\BIBentryALTinterwordspacing
H.~Touvron, M.~Cord, M.~Douze, F.~Massa, A.~Sablayrolles, and H.~Jégou, ``Training data-efficient image transformers \& distillation through attention,'' 2021. [Online]. Available: \url{https://arxiv.org/abs/2012.12877}
\BIBentrySTDinterwordspacing

\bibitem{Liu2021}
\BIBentryALTinterwordspacing
Z.~Liu, Y.~Lin, Y.~Cao, H.~Hu, Y.~Wei, Z.~Zhang, S.~Lin, and B.~Guo, ``Swin transformer: Hierarchical vision transformer using shifted windows,'' in \emph{2021 IEEE/CVF International Conference on Computer Vision (ICCV)}.\hskip 1em plus 0.5em minus 0.4em\relax Los Alamitos, CA, USA: IEEE Computer Society, Oct 2021, pp. 9992--10\,002. [Online]. Available: \url{https://doi.ieeecomputersociety.org/10.1109/ICCV48922.2021.00986}
\BIBentrySTDinterwordspacing

\bibitem{Guda2022}
\BIBentryALTinterwordspacing
V.~Guda, S.~Mugisha, C.~Chevallereau, M.~Zoppi, R.~Molfino, and D.~Chablat, ``Motion strategies for a cobot in a context of intermittent haptic interface,'' \emph{Journal of Mechanisms and Robotics}, vol.~14, no.~4, p. 041012, 06 2022. [Online]. Available: \url{https://doi.org/10.1115/1.4054509}
\BIBentrySTDinterwordspacing

\bibitem{han2021}
\BIBentryALTinterwordspacing
K.~Han, A.~Xiao, E.~Wu, J.~Guo, C.~Xu, and Y.~Wang, ``Transformer in transformer,'' 2021. [Online]. Available: \url{https://arxiv.org/abs/2103.00112}
\BIBentrySTDinterwordspacing

\bibitem{Howard2019}
\BIBentryALTinterwordspacing
A.~Howard, M.~Sandler, B.~Chen, W.~Wang, L.-C. Chen, M.~Tan, G.~Chu, V.~Vasudevan, Y.~Zhu, R.~Pang, H.~Adam, and Q.~Le, ``{ Searching for MobileNetV3 },'' in \emph{2019 IEEE/CVF International Conference on Computer Vision (ICCV)}.\hskip 1em plus 0.5em minus 0.4em\relax Los Alamitos, CA, USA: IEEE Computer Society, Nov. 2019, pp. 1314--1324. [Online]. Available: \url{https://doi.ieeecomputersociety.org/10.1109/ICCV.2019.00140}
\BIBentrySTDinterwordspacing

\bibitem{hinton2015}
\BIBentryALTinterwordspacing
G.~Hinton, O.~Vinyals, and J.~Dean, ``Distilling the knowledge in a neural network,'' 2015. [Online]. Available: \url{https://arxiv.org/abs/1503.02531}
\BIBentrySTDinterwordspacing

\bibitem{Guo2024}
\BIBentryALTinterwordspacing
Z.~Guo, D.~Wang, Q.~He, and P.~Zhang, ``Leveraging logit uncertainty for better knowledge distillation,'' \emph{Scientific Reports}, vol.~14, no.~1, p. 31249, Dec 2024. [Online]. Available: \url{https://doi.org/10.1038/s41598-024-82647-6}
\BIBentrySTDinterwordspacing

\bibitem{liu_et_al_2024}
B.~Liu, S.~Wei, F.~Zhang, N.~Guo, H.~Fan, and W.~Yao, ``Tomato leaf disease recognition based on multi-task distillation learning,'' \emph{FRONTIERS IN PLANT SCIENCE}, 2024.

\bibitem{yahya2023}
\BIBentryALTinterwordspacing
Y.~Alqahtani, M.~Nawaz, T.~Nazir, A.~Javed, F.~Jeribi, and A.~Tahir, ``An improved deep learning approach for localization and recognition of plant leaf diseases,'' \emph{Expert Systems with Applications}, vol. 230, p. 120717, 2023. [Online]. Available: \url{https://www.sciencedirect.com/science/article/pii/S0957417423012198}
\BIBentrySTDinterwordspacing

\bibitem{vaswani2023}
\BIBentryALTinterwordspacing
A.~Vaswani, N.~Shazeer, N.~Parmar, J.~Uszkoreit, L.~Jones, A.~N. Gomez, L.~Kaiser, and I.~Polosukhin, ``Attention is all you need,'' 2023. [Online]. Available: \url{https://arxiv.org/abs/1706.03762}
\BIBentrySTDinterwordspacing

\bibitem{zagoruyko2017}
\BIBentryALTinterwordspacing
S.~Zagoruyko and N.~Komodakis, ``Paying more attention to attention: Improving the performance of convolutional neural networks via attention transfer,'' 2017. [Online]. Available: \url{https://arxiv.org/abs/1612.03928}
\BIBentrySTDinterwordspacing

\bibitem{Yuan2020}
L.~Yuan, F.~E. Tay, G.~Li, T.~Wang, and J.~Feng, ``Revisiting knowledge distillation via label smoothing regularization,'' in \emph{2020 IEEE/CVF Conference on Computer Vision and Pattern Recognition (CVPR)}, 2020, pp. 3902--3910.

\bibitem{Vasisht2017farm}
D.~Vasisht, Z.~Kapetanovic, J.-h. Won, X.~Jin, R.~Chandra, A.~Kapoor, S.~N. Sinha, M.~Sudarshan, and S.~Stratman, ``Farmbeats: an iot platform for data-driven agriculture,'' in \emph{Proceedings of the 14th USENIX Conference on Networked Systems Design and Implementation}, ser. NSDI'17.\hskip 1em plus 0.5em minus 0.4em\relax USA: USENIX Association, 2017, p. 515–528.

\bibitem{hughes2016}
\BIBentryALTinterwordspacing
D.~P. Hughes and M.~Salathe, ``An open access repository of images on plant health to enable the development of mobile disease diagnostics,'' 2016. [Online]. Available: \url{https://arxiv.org/abs/1511.08060}
\BIBentrySTDinterwordspacing

\bibitem{Krizhevsky2017}
\BIBentryALTinterwordspacing
A.~Krizhevsky, I.~Sutskever, and G.~E. Hinton, ``Imagenet classification with deep convolutional neural networks,'' \emph{Commun. ACM}, vol.~60, no.~6, p. 84–90, May 2017. [Online]. Available: \url{https://doi.org/10.1145/3065386}
\BIBentrySTDinterwordspacing

\bibitem{simonyan2015}
\BIBentryALTinterwordspacing
K.~Simonyan and A.~Zisserman, ``Very deep convolutional networks for large-scale image recognition,'' 2015. [Online]. Available: \url{https://arxiv.org/abs/1409.1556}
\BIBentrySTDinterwordspacing

\bibitem{howard2017}
\BIBentryALTinterwordspacing
A.~G. Howard, M.~Zhu, B.~Chen, D.~Kalenichenko, W.~Wang, T.~Weyand, M.~Andreetto, and H.~Adam, ``Mobilenets: Efficient convolutional neural networks for mobile vision applications,'' 2017. [Online]. Available: \url{https://arxiv.org/abs/1704.04861}
\BIBentrySTDinterwordspacing

\bibitem{wu2020visualtransformer}
\BIBentryALTinterwordspacing
B.~Wu, C.~Xu, X.~Dai, A.~Wan, P.~Zhang, Z.~Yan, M.~Tomizuka, J.~Gonzalez, K.~Keutzer, and P.~Vajda, ``Visual transformers: Token-based image representation and processing for computer vision,'' 2020. [Online]. Available: \url{https://arxiv.org/abs/2006.03677}
\BIBentrySTDinterwordspacing

\bibitem{mehta2022}
\BIBentryALTinterwordspacing
S.~Mehta and M.~Rastegari, ``Mobilevit: Light-weight, general-purpose, and mobile-friendly vision transformer,'' 2022. [Online]. Available: \url{https://arxiv.org/abs/2110.02178}
\BIBentrySTDinterwordspacing

\bibitem{Al-Sharafi2022}
\BIBentryALTinterwordspacing
A.~Musa, M.~Hassan, M.~Hamada, and F.~Aliyu, ``Low-power deep learning model for plant disease detection for smart-hydroponics using knowledge distillation techniques,'' \emph{Journal of Low Power Electronics and Applications}, vol.~12, no.~2, 2022. [Online]. Available: \url{https://www.mdpi.com/2079-9268/12/2/24}
\BIBentrySTDinterwordspacing

\bibitem{tan2020efficient}
\BIBentryALTinterwordspacing
M.~Tan and Q.~V. Le, ``Efficientnet: Rethinking model scaling for convolutional neural networks,'' 2020. [Online]. Available: \url{https://arxiv.org/abs/1905.11946}
\BIBentrySTDinterwordspacing

\bibitem{jiang_et_al_2021}
Y.~Jiang, B.~Sharma, M.~Madhavi, and H.~Li, ``Knowledge distillation from bert transformer to speech transformer for intent classification,'' \emph{ARXIV-CS.CL}, 2021.

\bibitem{Li2023}
\BIBentryALTinterwordspacing
G.~Li, Y.~Wang, Q.~Zhao, P.~Yuan, and B.~Chang, ``Pmvt: a lightweight vision transformer for plant disease identification on mobile devices,'' \emph{Frontiers in Plant Science}, vol.~14, 2023. [Online]. Available: \url{https://www.frontiersin.org/journals/plant-science/articles/10.3389/fpls.2023.1256773}
\BIBentrySTDinterwordspacing

\bibitem{ji2021showattenddistillknowledgedistillation}
\BIBentryALTinterwordspacing
M.~Ji, B.~Heo, and S.~Park, ``Show, attend and distill:knowledge distillation via attention-based feature matching,'' 2021. [Online]. Available: \url{https://arxiv.org/abs/2102.02973}
\BIBentrySTDinterwordspacing

\bibitem{Shen2024}
C.~Pham, V.-A. Nguyen, T.~Le, D.~Phung, G.~Carneiro, and T.-T. Do, ``Frequency attention for knowledge distillation,'' in \emph{2024 IEEE/CVF Winter Conference on Applications of Computer Vision (WACV)}, 2024, pp. 2266--2275.

\bibitem{Yim2017}
J.~Yim, D.~Joo, J.~Bae, and J.~Kim, ``A gift from knowledge distillation: Fast optimization, network minimization and transfer learning,'' in \emph{2017 IEEE Conference on Computer Vision and Pattern Recognition (CVPR)}, 2017, pp. 7130--7138.

\bibitem{habib2024knowledgedistillationvit}
\BIBentryALTinterwordspacing
G.~Habib, T.~J. Saleem, and B.~Lall, ``Knowledge distillation in vision transformers: A critical review,'' 2024. [Online]. Available: \url{https://arxiv.org/abs/2302.02108}
\BIBentrySTDinterwordspacing

\bibitem{liu2024cross}
\BIBentryALTinterwordspacing
Y.~Liu, J.~Cao, B.~Li, W.~Hu, J.~Ding, L.~Li, and S.~Maybank, ``Cross-architecture knowledge distillation,'' \emph{Int. J. Comput. Vision}, vol. 132, no.~8, p. 2798–2824, Feb. 2024. [Online]. Available: \url{https://doi.org/10.1007/s11263-024-02002-0}
\BIBentrySTDinterwordspacing

\bibitem{Yang2024ViTKD}
Z.~Yang, Z.~Li, A.~Zeng, Z.~Li, C.~Yuan, and Y.~Li, ``Vitkd: Feature-based knowledge distillation for vision transformers,'' in \emph{2024 IEEE/CVF Conference on Computer Vision and Pattern Recognition Workshops (CVPRW)}, 2024, pp. 1379--1388.

\bibitem{TZOUNIS2017}
\BIBentryALTinterwordspacing
A.~Tzounis, N.~Katsoulas, T.~Bartzanas, and C.~Kittas, ``Internet of things in agriculture, recent advances and future challenges,'' \emph{Biosystems Engineering}, vol. 164, pp. 31--48, 2017. [Online]. Available: \url{https://www.sciencedirect.com/science/article/pii/S1537511017302544}
\BIBentrySTDinterwordspacing

\bibitem{Farooq2020}
\BIBentryALTinterwordspacing
M.~S. Farooq, S.~Riaz, A.~Abid, T.~Umer, and Y.~B. Zikria, ``Role of iot technology in agriculture: A systematic literature review,'' \emph{Electronics}, vol.~9, no.~2, 2020. [Online]. Available: \url{https://www.mdpi.com/2079-9292/9/2/319}
\BIBentrySTDinterwordspacing

\bibitem{Qu2020H-AT}
Y.~Qu, W.~Deng, and J.~Hu, ``H-at: Hybrid attention transfer for knowledge distillation,'' in \emph{Pattern Recognition and Computer Vision}, Y.~Peng, Q.~Liu, H.~Lu, Z.~Sun, C.~Liu, X.~Chen, H.~Zha, and J.~Yang, Eds.\hskip 1em plus 0.5em minus 0.4em\relax Cham: Springer International Publishing, 2020, pp. 249--260.

\bibitem{Gehlot_2023_tomato_village}
\BIBentryALTinterwordspacing
M.~Gehlot, R.~K. Saxena, and G.~C. Gandhi, ``“tomato-village”: a dataset for end-to-end tomato disease detection in a real-world environment,'' \emph{Multimedia Syst.}, vol.~29, no.~6, p. 3305–3328, Aug. 2023. [Online]. Available: \url{https://doi.org/10.1007/s00530-023-01158-y}
\BIBentrySTDinterwordspacing

\bibitem{Huang_2020_tomato_disease}
\BIBentryALTinterwordspacing
Y.-H. Huang, Mei-Ling;~Chang, ``Dataset of tomato leaves,'' \emph{Mendeley Data}, vol.~V1, 2020. [Online]. Available: \url{https://data.mendeley.com/datasets/ngdgg79rzb/1}
\BIBentrySTDinterwordspacing

\bibitem{Tan2019}
\BIBentryALTinterwordspacing
M.~Tan and Q.~Le, ``{E}fficient{N}et: Rethinking model scaling for convolutional neural networks,'' in \emph{Proceedings of the 36th International Conference on Machine Learning}, ser. Proceedings of Machine Learning Research, K.~Chaudhuri and R.~Salakhutdinov, Eds., vol.~97.\hskip 1em plus 0.5em minus 0.4em\relax PMLR, 09--15 Jun 2019, pp. 6105--6114. [Online]. Available: \url{https://proceedings.mlr.press/v97/tan19a.html}
\BIBentrySTDinterwordspacing

\bibitem{Mark2018}
\BIBentryALTinterwordspacing
M.~Sandler, A.~G. Howard, M.~Zhu, A.~Zhmoginov, and L.~Chen, ``Inverted residuals and linear bottlenecks: Mobile networks for classification, detection and segmentation,'' \emph{CoRR}, vol. abs/1801.04381, 2018. [Online]. Available: \url{http://arxiv.org/abs/1801.04381}
\BIBentrySTDinterwordspacing

\bibitem{Lin_2023}
\BIBentryALTinterwordspacing
J.~Lin, L.~Zhu, W.-M. Chen, W.-C. Wang, and S.~Han, ``Tiny machine learning: Progress and futures [feature],'' \emph{IEEE Circuits and Systems Magazine}, vol.~23, no.~3, p. 8–34, 2023. [Online]. Available: \url{http://dx.doi.org/10.1109/MCAS.2023.3302182}
\BIBentrySTDinterwordspacing

\end{thebibliography}
\vspace{11pt}

\begin{IEEEbiography}[{\includegraphics[width=1in,height=1.25in,clip,keepaspectratio]{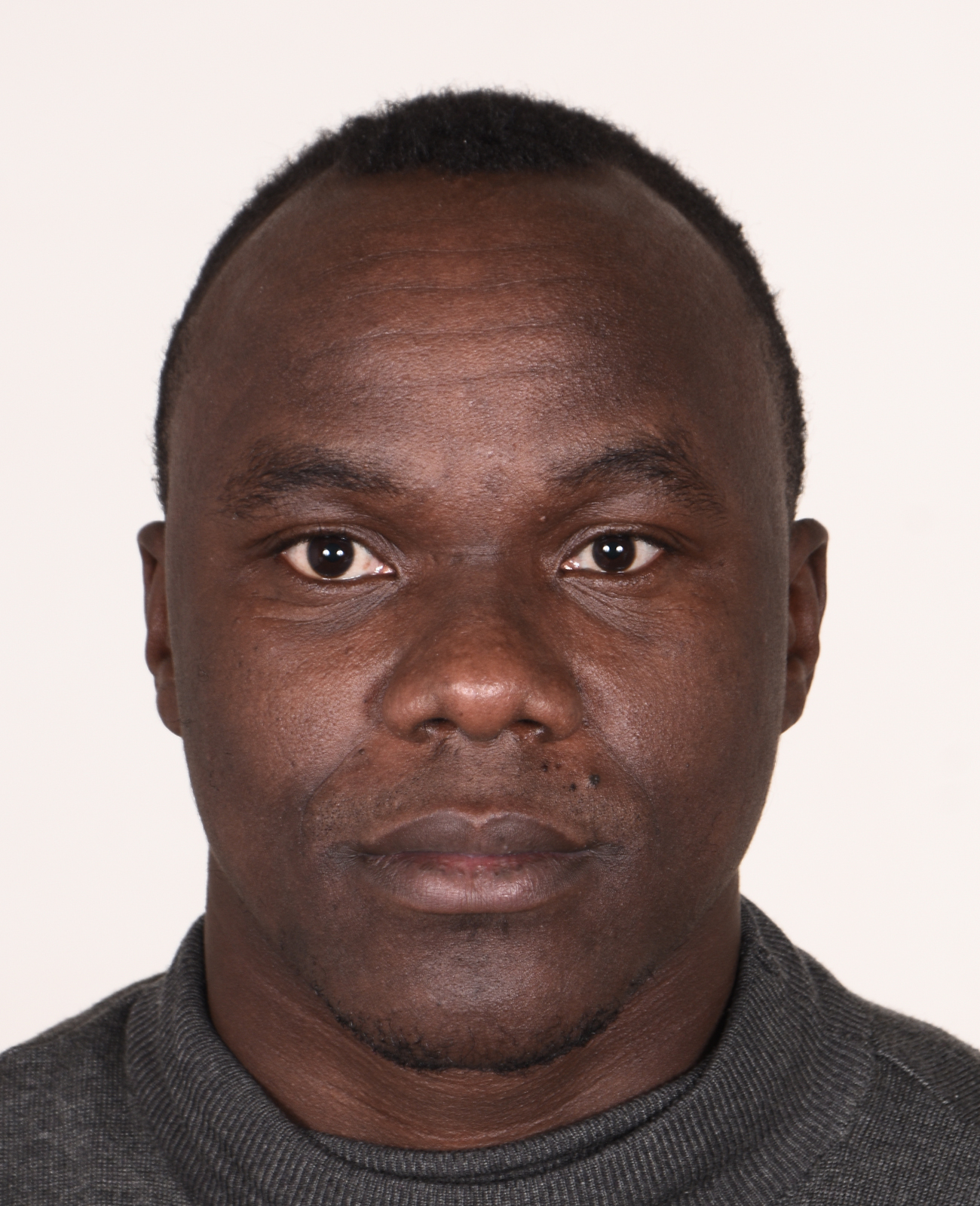}}]{Stanley Mugisha}
received the Bsc. and Msc. degrees in Computer science from the University of Mysore India, in 2017 and the Ph.D. degree in robotics and mechatronics from University of Genova Italy and Ecole Centrale de Nantes, France, in 2022.
In 2023, He worked as a postdoc research fellow in human robot collaboration at the university of Tartu in Estonia. Since 2024, he works at Soroti university as a lecturer. His research interest include: Computer vision, human robot interaction,  IoT, and embedded Machine Learning.
\end{IEEEbiography}

\begin{IEEEbiographynophoto}{Rashid Kisitu}
received his B.Eng in Electronics and Computer engineering from Soroti University in 2025. Currently he is working as a research assistant in the department of electronics and computer engineerig, school of engineering and technology at Soroti university. He has worked on various projects in computer vision and IoT for maternal health. His research interests include machine learning, embedded systems, computer vision and IoT.
\end{IEEEbiographynophoto}

\begin{IEEEbiographynophoto}{Florence Tushabe}
received her BSc in Computer Science in 2000 from University of Dar-es-salaam, Tanzania,  MSc. Computer Science from Makerere University in 2004 and PhD Computer Science, University of Groningen, Netherlands in 2010. From 2011-2018,  She worked with Makerere University as a senior lecturer and head of department Computer science in the school of computing and information sciences, Since 2019, She currently working as an Associate Professor at Soroti University in the school of Engineering and technology and department of electronics and computer engineering . She is a distinguished researcher and has authored three books and more than 50 articles. She is passionate about AI (Artificial Intelligence) and has developed intelligent systems for object recognition in images and speech-to-text recognition. Her research interests include: Computer vision , intelligent systems, audio and speech processing and NLP.
\end{IEEEbiographynophoto}

\vfill

\end{document}